\documentclass{article}

\usepackage{microtype}
\usepackage{graphicx}
\usepackage{subcaption}
\usepackage{booktabs} 

\usepackage{hyperref}

\usepackage{booktabs}
\usepackage{amssymb}
\usepackage{multirow}

\usepackage[accepted]{icml2026}

\usepackage{amsmath}
\usepackage{amssymb}
\usepackage{mathtools}
\usepackage{amsthm}

\usepackage[capitalize,noabbrev]{cleveref}

\theoremstyle{plain}

\theoremstyle{definition}

\theoremstyle{remark}

\usepackage[textsize=tiny]{todonotes}

\icmltitlerunning{Autoregressive EHR Foundation Models with Multimodal Inputs}

\begin{document}

\twocolumn[
  \icmltitle{Autoregressive EHR Foundation Models with Multimodal Inputs}



  \icmlsetsymbol{equal}{*}
  
  \begin{icmlauthorlist}
    \icmlauthor{Yuxuan Liu}{equal,ic_computing,ai4health}
    \icmlauthor{Joshua Placidi}{equal,ic_computing,ai4health}
    \icmlauthor{Jinpei Han}{equal,ic_computing}
    \icmlauthor{Alfred John Balston}{ic_computing,ai4health,public_heatlh}
    \icmlauthor{Marek Rei}{ic_computing}
    \icmlauthor{A. Aldo Faisal}{ic_computing,ai4health,ic_bioeng,bayreuth}
  \end{icmlauthorlist}

  \icmlaffiliation{ic_computing}{Dept. of Computing, Imperial College London, London, United Kingdom}
  \icmlaffiliation{ic_bioeng}{Dept. of Bioengineering, Imperial College London, London, United Kingdom}
  \icmlaffiliation{ai4health}{UKRI Centres in AI for Health, United Kingdom}
  \icmlaffiliation{public_heatlh}{School of Public Health, Imperial College London, London, United Kingdom}
  \icmlaffiliation{bayreuth}{Chair in Digital Health, Universität  Bayreuth, Bayreuth, Germany}
  
  \icmlcorrespondingauthor{Joshua Placidi}{joshua.placidi25@imperial.ac.uk}

  \icmlcorrespondingauthor{Yuxuan Liu}{edison.liu22@imperial.ac.uk}
  \icmlcorrespondingauthor{A. Aldo Faisal}{aldo.faisal@imperial.ac.uk}

  \icmlkeywords{Electronic Health Records, Foundation Models, Multimodal Learning, Time Series}

  \vskip 0.3in
]



\printAffiliationsAndNotice{\footnotesize *Equal contribution.}  

\begin{abstract}
Autoregressive foundation models trained on tokenized electronic health records (EHRs) can support zero-shot clinical prediction, yet most operate on structured event codes alone, and do not incorporate multiple modalities in a principled way.
We present a framework for conditioning such models on auxiliary clinical modalities, including ECG waveforms, chest X-ray images, and clinical notes, using modality-specific latent compression and gated cross-attention with temporal alignment.
We investigate two key design choices:
(1) how to compress long per-modality sequences (e.g., ECG time series) before they enter the multi-modal cross-attention. This feature may be essential to reduce compute overheads and may be beneficial for generalization;
(2) how the choice of pretrained encoder for each modality impacts downstream performance.
Through controlled ablations on MIMIC-IV, we show that the best latent-compression configurations outperforms both uncompressed cross-attention and mean pooling. Encoder choice has a clear within-modality effect, with stronger pretrained encoders consistently outperforming weaker alternatives. We further show that merely adding auxiliary modalities does not guarantee improvement on ICU mortality prediction over an EHR-only baseline. This implies that careful design of the fusion architecture and an appropriate evaluation in the clinical context are required.
\end{abstract}

\begin{figure}[t]
    \centering
    \includegraphics[width=0.95\columnwidth]{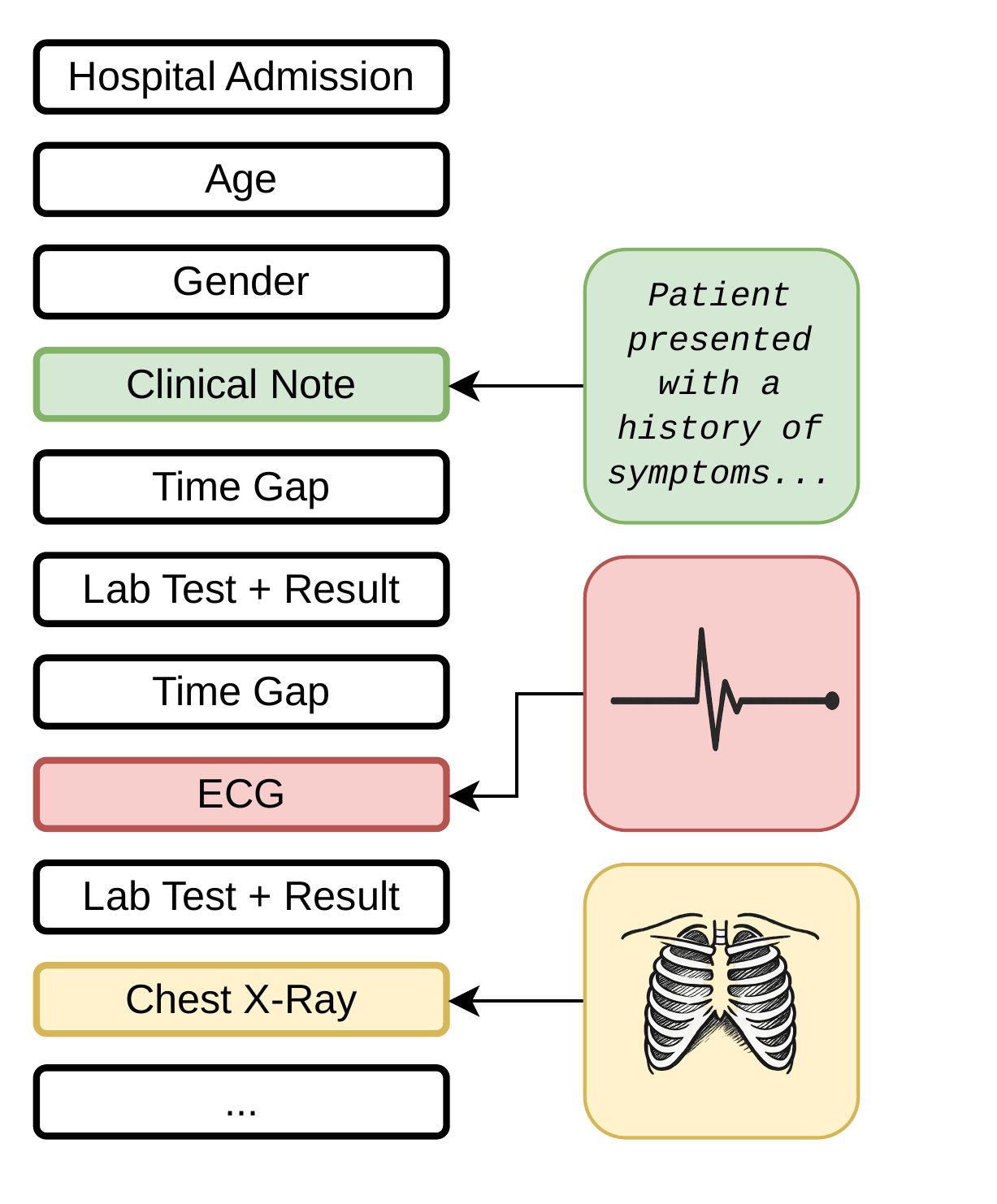}
    \caption{Electronic health records are represented as sequences of clinical events spanning multiple modalities.}
    \label{fig:multimodal_ehr}
\end{figure}

\begin{figure*}[t]
\centering
\includegraphics[width=0.95\textwidth]{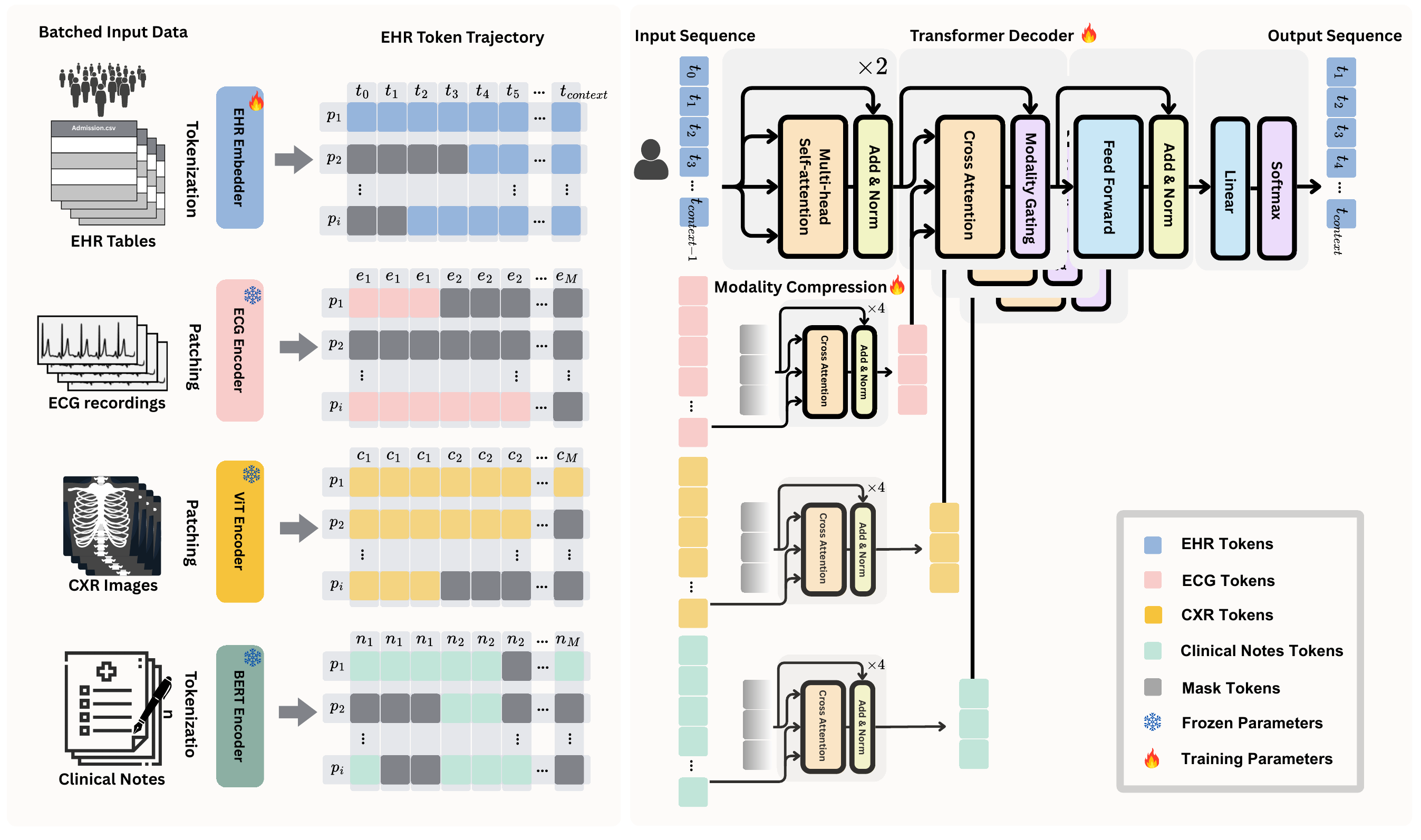}
\caption{Multimodal EHR trajectory model. Auxiliary modalities (ECG, CXR, notes) are encoded, compressed into latent tokens, and fused into a GPT-2-style decoder via gated cross-attention. Temporal masking ensures each EHR token attends only to past modality events.}
\label{fig:framework}
\end{figure*}

\section{Introduction}
Recent work has shown that autoregressive transformers trained on tokenized electronic health records (EHRs) can learn reusable patient representations that support zero-shot clinical prediction \cite{renc2024ethos, waxler2025generative}. These models linearize longitudinal EHR events (i.e. diagnoses, procedures, medications, vitals) into discrete token sequences and apply next-token prediction, analogous to language modeling \cite{pang2021cehr}. However, most EHR foundation models operate on a single modality of structured event codes, even though clinical decision-making routinely draws on complementary multimodal evidence as shown in Fig. \ref{fig:multimodal_ehr}. Electrocardiograms (ECG) provide direct measurements of cardiac electrical activity and support the detection of arrhythmias, conduction abnormalities, and ischemic patterns \cite{gu2025foundation}. Chest X-ray (CXR) images offer rapid evidence of pulmonary and cardiac processes such as consolidation, edema, cardiomegaly, and pleural effusions. Clinical notes capture information that is not well represented in coded tables, including presenting symptoms, differential diagnoses, and clinician assessments \cite{wang2024multimodal}. Integrating these modalities into a generative EHR model is nontrivial.

Each modality differs in dimensionality, sampling rate, and temporal availability; missingness is clinically informative rather than random; and naive fusion risks temporal leakage when auxiliary observations acquired after an EHR event are allowed to influence its prediction \cite{spathis2024first}. Prior multimodal clinical work has largely adopted late fusion---aggregating predictions from independently trained unimodal models---which cannot capture fine-grained cross-modal dependencies \cite{stahlschmidt2022multimodal, jorf2025medpatch}. Intermediate fusion via cross-attention, as popularized by Flamingo \cite{alayrac2022flamingo} and adapted for medical tasks \cite{moor2023med, tu2024towards, amar2025integrating}, offers a more expressive alternative but has not been systematically evaluated for autoregressive EHR trajectory modeling.

In this work, we augment a decoder-only EHR backbone to use ECG, CXR, and clinical notes with latent compression and  cross-attention.
We propose a compression module, inspired by Perceiver \cite{jaegle2021perceiver}, that compresses modality information into a small number of latents, improving model efficiency and performance.
Through controlled ablations on MIMIC-IV, we show that latent compression is essential for effective and efficient fusion of long modality sequences, and that encoder choice has a clear within-modality effect. However, we also find that simply adding auxiliary modalities does not guarantee improvement on ICU mortality: only one bimodal configuration (notes with BioMedBERT) marginally exceeds the EHR-only baseline, and a unified three-modality model does not surpass it either. This points to a need for more flexible fusion architectures and clinically contextual evaluation that can reveal where each modality genuinely contributes.

\section{Method}

\subsection{Problem Setup and Data}
We study autoregressive modeling of patient trajectories from MIMIC-IV v3.1 \cite{johnson2024mimiciv31}. Following ETHOS \cite{renc2024ethos}, we convert MIMIC-IV and MIMIC-IV-ED \cite{johnson2023mimicived22} into the Medical Event Data Standard (MEDS) format \cite{arnrich2024meds} and tokenize each patient's record into a chronologically ordered sequence $x_i = (x_{i,1}, \dots, x_{i,T_i})$ over a vocabulary $V$ of medical-event tokens, with event timestamps $\tau_{i,t}$.

In addition to the EHR stream, each patient $p_i$ may have auxiliary observations from a modality set $\mathcal{M}=\{\mathrm{ecg},\mathrm{cxr},\mathrm{note}\}$, with available modalities $\mathcal{M}_i \subseteq \mathcal{M}$ varying across patients. Specifically, patient $p_i$ may have ECG recordings $E_i=\{(e_{i,r},\, \tau^{\mathrm{ecg}}_{i,r})\}_{r=1}^{R_i}$, where $e_{i,r}\in\mathbb{R}^{12\times L_{i,r}}$ is a 12-lead waveform and $\tau^{\mathrm{ecg}}_{i,r}$ is the acquisition time; chest radiographs $C_i=\{(c_{i,j},\, \tau^{\mathrm{cxr}}_{i,j})\}_{j=1}^{J_i}$; and clinical notes $N_i=\{(n_{i,k},\, \tau^{\mathrm{note}}_{i,k})\}_{k=1}^{K_i}$, where each note $n_{i,k}$ is tokenized into a sequence $u_{i,k}=(u_{i,k,1},\dots,u_{i,k,L_{i,k}})$ and truncated or padded to the encoder context length $L_{\max}$. All auxiliary data are sourced from MIMIC-IV-ECG \cite{gow2023mimic}, MIMIC-CXR-JPG \cite{johnson2019mimic}, and MIMIC-IV-Note \cite{johnson2023mimic} via PhysioNet \cite{goldberger2000physionet}. Missing modalities are represented via explicit masks rather than imputation. We use an 80/10/10 subject-level train/validation/test split.

\subsection{Multimodal Architecture}
Our Multimodal EHR trajectory model (see Fig.~2) reads an EHR  and auxiliary modalities (here ECG time series, CXR images, and clinical notes text) token sequence that are embedded via a learned lookup table and processed by a GPT-2-style decoder \cite{radford2019gpt2} with latent dimension $D$.
Each auxiliary modality is encoded by a frozen pretrained, modality-specific encoder.
For ECG, each recording is partitioned into contiguous temporal patches and projected into patch embeddings.
For CXR, images are split into 2D patches following the standard vision-transformer approach.
For clinical notes, each document is tokenized and encoded by a BERT-style model, with truncation or padding to a fixed context length.

\paragraph{Latent compression.}
Each modality encoder produces a sequence of embeddings whose length depends on the input (e.g., a clinical note of $N$ tokens yields an $(N, d_{\textrm{encoder}})$ tensor, where $d_{\textrm{encoder}}$ is the encoder output dimension).
Naive cross-attention from the EHR sequence of length $T$ to such a modality sequence requires $\mathcal{O}(NT)$ operations, which quickly becomes expensive.
We propose a latent compression module that compresses each modality sequence into $k$ latents with $k \ll N$.
The latent queries are learned parameters initialized randomly and iteratively refined through $l$ layers of cross-attention to the modality sequence, reducing the effective cross-attention cost to $\mathcal{O}(Nk+Tk)$.


\paragraph{Gated cross-attention.}
The decoder conditions on compressed modality latents via cross-attention blocks interleaved with self-attention layers. Following Flamingo \cite{alayrac2022flamingo}, each cross-attention block is modulated by a learnable scalar gate $g^l = \sigma(\beta^l)$ that controls the contribution of the modality update at layer $l$, stabilizing training and enabling depth-dependent modality usage.

\paragraph{Temporal alignment masking.}
To ensure that predictions do not condition on future auxiliary observations, we enforce a past-only cross-attention rule: an EHR token at time $\tau_t$ may only attend to modality events with acquisition time strictly before $\tau_t$ (equal timestamps are also masked). Modality instances are retrieved by subject ID within the EHR window's time span, with at most $K_m$ events per modality. This causal mask is combined with event-availability and padding masks to handle partial or absent modality availability.

\begin{table}
\centering
\small
\setlength{\tabcolsep}{6pt}
\begin{tabular}{lll|ccc}
\toprule
 &  &  & \multicolumn{3}{c}{ICU Mortality} \\
\cmidrule(lr){4-6}
Compression & $k$ & $l$ & AUROC & AUPRC & F1 \\
\midrule
None & -- & -- & 0.8565 & 0.4357 & 0.7061 \\
\midrule
Mean Pooling & -- & -- & 0.8642 & 0.4274 & 0.7037 \\
\midrule
Latent & 8  & 1 & 0.8662 & 0.4400 & 0.7082 \\
Latent & 8  & 2 & 0.8709 & 0.4444 & \textbf{0.7131} \\
Latent & 8  & 4 & 0.8751 & \textbf{0.4599} & 0.7107 \\
Latent & 16 & 1 & 0.8640 & 0.4426 & 0.7121 \\
Latent & 16 & 2 & \textbf{0.8771} & 0.4486 & 0.7082 \\
Latent & 16 & 4 & 0.8580 & 0.4235 & 0.6957 \\
\bottomrule
\end{tabular}
\caption{Analysis of different compression methods in a bimodal model trained with EHR and clinical notes. Clinical notes were encoded using BioMedBERT \cite{pubmedbert} as the note encoder. Performance is reported on the ICU Mortality task.}
\label{tab:latent_compression}
\end{table}

\paragraph{Training objective.}
We train with standard next-token cross-entropy loss on the EHR token stream: $\mathcal{L}_{\mathrm{CE}}(\theta) = - \sum_{u} \log p_\theta(x_{i,u+1} \mid x_{i,\leq u},\, E_i, C_i, N_i)$, where auxiliary modalities serve only as conditioning inputs and the model generates EHR tokens only. The loss excludes padded and demographic positions.

\begin{table*}[t]
\centering
\caption{Effect of modality and encoder choice on ICU mortality. Bimodal rows pair EHR with one auxiliary modality; the final row reports a unified model conditioned on all three modalities using the best-performing encoder per modality (BioMedCLIP, CSFM, BioMedBERT). All multimodal variants use $k=8$ and $l=4$.}
\label{tab:modality_encoder_results}
\small
\setlength{\tabcolsep}{6pt}
\begin{tabular}{cccc|l|ccc}
\toprule
\multicolumn{4}{c|}{Modalities} & \multirow{2}{*}{Modality Encoder} & \multicolumn{3}{c}{ICU Mortality} \\
\cmidrule(lr){1-4} \cmidrule(lr){6-8}
EHR & CXR & ECG & Notes &  & AUROC & AUPRC & F1 \\
\midrule
$\checkmark$ & $\times$ & $\times$ & $\times$ & -            & 0.8651          & 0.4561          & 0.7182 \\
\midrule
$\checkmark$ & $\checkmark$ & $\times$ & $\times$ & BioMedCLIP ViT \cite{zhang2023biomedclip} & \textbf{0.8624}          & \textbf{0.4471}          & \textbf{0.7075} \\
$\checkmark$ & $\checkmark$ & $\times$ & $\times$ & ViT-MAE \cite{he2022mae}       & 0.8505          & 0.4315          & 0.6755 \\
\midrule
$\checkmark$ & $\times$ & $\checkmark$ & $\times$ & CSFM \cite{gu2026cardiac}          & \textbf{0.8703}          & \textbf{0.4263}          & \textbf{0.7024} \\
$\checkmark$ & $\times$ & $\checkmark$ & $\times$ & ECG-FM \cite{mckeen2025ecg}        & 0.8680          & 0.4072          & 0.7022 \\
\midrule
$\checkmark$ & $\times$ & $\times$ & $\checkmark$ & BERT \cite{devlin2019bert}         & 0.8579          & 0.4164          & 0.7066 \\
$\checkmark$ & $\times$ & $\times$ & $\checkmark$ & BioMedBERT \cite{pubmedbert}   & \textbf{0.8751} & \textbf{0.4599} & \textbf{0.7107} \\
\midrule
$\checkmark$ & $\checkmark$ & $\checkmark$ & $\checkmark$ & Best per modality & 0.8460 & 0.4330 & 0.6997 \\
\bottomrule
\end{tabular}
\end{table*}

\subsection{Modality Encoders}
We evaluate multiple pretrained encoders per modality to assess the impact of encoder choice on downstream performance, while keeping the EHR backbone and fusion mechanism fixed. For \textbf{ECG}, we compare CSFM \cite{gu2026cardiac}, a multi-dataset cardiac foundation model, with ECG-FM \cite{mckeen2025ecg}, an open electrocardiogram foundation model; both produce patch-level embeddings from 12-lead waveforms. For \textbf{CXR}, we compare BioMedCLIP, a vision--language model pretrained on biomedical image--text pairs, with a ViT-MAE pretrained via masked autoencoding on natural images. For \textbf{clinical notes}, we compare BERT \cite{devlin2019bert} with BioMedBERT \cite{pubmedbert}, a domain-adapted variant pretrained on PubMed abstracts. All encoders are frozen during training; the latent compression module maps their outputs into $k$ latent vectors of dimension $D$, matching the backbone hidden dimension.

\subsection{Zero-Shot Evaluation}
We focus on zero-shot ICU mortality prediction. Following ETHOS \cite{renc2024ethos}, evaluation conditions on the last $L$ tokens of patient history ending at the task start event (e.g., the ICU admission token for ICU mortality) and samples $K{=}20$ stochastic futures; the predicted probability is the fraction of rollouts containing the target token within the task horizon. For multimodal models, auxiliary encodings are computed once from the observed window and kept fixed during rollout.
Further details of the rollout-based evaluation pipeline, including termination criteria and metric computation, are provided in Appendix~\ref{app:framework}.

\section{Results \& Discussion}
Table~\ref{tab:latent_compression} reports a compression ablation in a bimodal EHR + clinical notes model on ICU mortality. Mean pooling improves AUROC over no compression but slightly lowers AUPRC and F1, whereas best latent compression configuration improves all three. Sweeping $k \in \{8, 16\}$ and $l \in \{1, 2, 4\}$, we select $k{=}8, l{=}4$, which gives the best AUPRC (0.4599) and competitive AUROC; we adopt it for all subsequent experiments. We note that no single setting dominates across all three metrics, so this choice prioritises AUPRC given the class imbalance of the mortality task.

\paragraph{Multimodal encoder ablation.}
Table~\ref{tab:modality_encoder_results} evaluates pretrained encoder choice in bimodal settings on ICU mortality, with the EHR-only baseline at 0.8651/0.4561/0.7182 (AUROC/AUPRC/F1). Adding an auxiliary modality does not uniformly improve over this baseline: CXR with either encoder performs below the baseline across all three metrics; ECG with CSFM \cite{gu2026cardiac} improves AUROC (0.8703 vs.\ 0.8651) but lowers AUPRC and F1; clinical notes with BioMedBERT \cite{pubmedbert} is the only configuration to exceed the baseline on both AUROC and AUPRC (0.8751/0.4599), while still trailing on F1. Encoder choice nonetheless has a clear within-modality effect: BioMedCLIP \cite{zhang2023biomedclip} outperforms ViT-MAE \cite{he2022mae} on CXR; CSFM outperforms ECG-FM \cite{mckeen2025ecg} on ECG with a notable AUPRC gap (0.426 vs.\ 0.407); and BioMedBERT substantially outperforms BERT \cite{devlin2019bert} on notes across all metrics. For CXR and notes, the advantage comes from domain-adapted pretraining over general-purpose models; for ECG, CSFM benefits from training on a larger and more diverse cardiac corpus. The unified model conditioned on all three modalities (Table~\ref{tab:modality_encoder_results}, final row; BioMedCLIP for CXR, CSFM for ECG, and BioMedBERT for notes) also fails to surpass the EHR-only baseline, indicating that simply combining modalities does not yield additive gains.

\paragraph{Discussion.}
Our results show that these design choices matter substantially, yet they are often left uninvestigated. 
We note, however, that the rollout-based evaluation is stochastic and can be noisy, and due to computational constraints we were unable to repeat all experiments across multiple random seeds.
These results should therefore be interpreted as evidence of broad trends rather than definitive rankings between closely performing configurations.
Three takeaways emerge from these results. First, learned latent compression setting substantially outperforms both uncompressed cross-attention and mean pooling (Table~\ref{tab:latent_compression}), while reducing the attention cost from $\mathcal{O}(NT)$ to $\mathcal{O}(Nk+Tk)$. This is consistent with the Perceiver and Flamingo line of work \cite{jaegle2021perceiver, alayrac2022flamingo}, in which a small set of learned latents acts as a bottleneck that summarises a long modality sequence: here the bottleneck appears to distil the clinically informative content of each modality more effectively than either attending over the full sequence or naively averaging it, with the added benefit of bounded compute regardless of modality sequence length.

Second, encoder choice has a clear within-modality effect (Table~\ref{tab:modality_encoder_results}): for every modality, the domain-adapted or more broadly pretrained encoder outperforms the general-purpose alternative, holding the EHR backbone and fusion mechanism fixed. This indicates that, in our setting, the quality of the frozen modality representation is at least as consequential as the fusion architecture, echoing observations from medical foundation-model work that domain-specific pretraining materially improves transfer \cite{pubmedbert, zhang2023biomedclip}. A practical implication of this finding is that effort spent selecting or adapting the modality encoder may yield larger returns than additional fusion-side complexity.

Third, and most importantly, adding modalities does not guarantee improvement on ICU mortality: only notes with BioMedBERT exceed the baseline on AUROC and AUPRC, and the unified three-modality model underperforms the EHR-only baseline. This is not unique to clinical data -- a consistent finding across multimodal learning is that the best unimodal network can match or beat a jointly trained multimodal one, because joint optimisation tends to be \emph{greedy}---the model comes to rely on whichever modality is fastest to learn from, leaving the others underfit \cite{gradientblending2020, wu2022greedy}, a phenomenon also framed as \emph{modality competition} under joint training \cite{huang2022modalitycompetition}. In our case the EHR token stream is a strong, dense, and temporally complete signal for mortality, whereas the auxiliary modalities are sparse, intermittently available, and masked to past-only events; under a single next-token objective the backbone has little incentive to exploit them. This reframes our negative result: the deficit is plausibly an optimisation and signal-balance problem rather than evidence that the modalities lack information.

The CXR result deserves an extra comment, as it is a negative result beyond non-additivity; it falls below EHR-only baseline on all metrics (AUROC, AUPRC, F1) with both evaluated encoders. CXR appears to be the temporally sparsest modality in our training set. We attribute this primarily to temporal availability: radiographs are sparse relative to the dense EHR window, and after past-only masking many prediction points have no admissible CXR study to attend to, so the cross-attention pathway contributes mostly noise while still consuming model capacity. With our data we cannot fully separate this limited availability from low intrinsic mortality signal in CXR. ECG, in contrast, though also sparse, carries a more directly mortality-relevant physiological signal and at least improves AUROC.

\section{Conclusion}

We presented a framework for conditioning autoregressive EHR foundation models on ECG, CXR, and clinical notes via latent compression and gated cross-attention. Controlled ablations on MIMIC-IV show that latent compression outperforms simpler aggregation while also improving efficiency ($\mathcal{O}(Nk+Tk)$ attention cost), encoder choice has a clear within-modality effect, and simply adding modalities does not consistently improve ICU mortality over a strong EHR-only baseline. Future work includes more flexible fusion and modality-aware training, and cohort-stratified evaluation that can reveal where each modality contributes most.


\section*{Acknowledgements}
YL is supported by UK Research and Innovation (UKRI) Centre for Doctoral Training in AI for Healthcare grant number (EP/S023283/1); 
JP is supported by the UKRI AI Centre for Doctoral Training in Digital Healthcare grant number (EP/Y030974/1), and by a philanthropic Bupa PhD Scholarship.
AJB is supported by the UKRI AI Centre for Doctoral Training in Digital Healthcare grant number (EP/Y030974/1), and by The Lee Family Faculty of Medicine Scholarship.
AAF is supported by the UKRI Turing AI Fellowship grant number (EP/V025449/1).
AAF and MR acknowledge support of the UKRI AI programme and the Engineering and Physical Sciences Research Council (EPSRC) for the AI Hub in Generative Models 
(EP/Y028805/1).
AAF \& MR acknowledge the use of resources provided by the Isambard-AI National AI Research Resource (AIRR).
Isambard-AI is operated by the University of Bristol and is funded by the UK Government’s Department for Science, Innovation and Technology (DSIT) via UK Research and Innovation; and the Science and Technology Facilities Council (ST/AIRR/I-A-I/1023).
\pagebreak

\bibliography{references}
\bibliographystyle{icml2026}

\appendix
\onecolumn

\section{Model, Training, and Evaluation Framework}
\label{app:framework}

\paragraph{Model.}
We use a GPT-2 style architecture.
The model consists of a learned token embedding layer, a learned positional embedding layer, a stack of transformer blocks with causal self-attention and MLP sublayers, a final layer normalization, and a linear language-modeling head.
The model is trained with a standard next-token prediction objective over tokenized EHR trajectories.
A ``reduce learning rate on plateau" scheduler is used to multiply the learning rate by $0.5$ when the validation loss plateaus for more than 3 validation intervals in a row. 
Table~\ref{tab:core_hparams} summarizes the core model and optimization hyperparameters used in our experiments.

\begin{table}[t]
\centering
\small
\setlength{\tabcolsep}{5pt}
\begin{tabular}{ll}
\toprule
\textbf{Component} & \textbf{Setting} \\
\midrule
Embedding dimension & 768 \\
Transformer layers & 6 (unless stated otherwise) \\
Attention heads & 12 \\
Context length & 2048 \\
Dropout & 0.1 \\
Batch size & 128 \\
Optimizer & AdamW \\
Learning rate & $3 \times 10^{-4}$ \\
Weight decay & 0.1 \\
Validation interval & every 10K steps \\
\bottomrule
\end{tabular}
\caption{Model and training hyperparameters used in our experiments.}
\label{tab:core_hparams}
\end{table}

\paragraph{Evaluation framework.}
 Evaluation is performed using a rollout-based benchmark framework as proposed in \cite{renc2024ethos}.
 For the ICU Mortality tasks we prompt the model with patient trajectories up to and including the presence of an \texttt{ICU\_ADMISSION} token.
 The model then generates the next token auto-regressively until it predicts either the \texttt{MEDS\_DEATH} or \texttt{ICU\_DISCHARGE} tokens, or until a maximum generation length of 4096 tokens is reached. For each prompt, we run 20 stochastic rollouts. A scalar score is computed as the fraction of rollouts that terminate with the different end tokens, and this score is used as the model output for binary classification. Simulations that exceed the 4096-token limit are terminated and excluded from benchmark calculations. AUROC and AUPRC are then computed from the resulting per-prompt scores.

%
%

\end{document}